\documentclass[conference]{IEEEtran}

\usepackage{xcolor}
\usepackage{graphicx}

\usepackage{amsmath} 

\usepackage[hyphens]{url}
\usepackage{hyperref}
\hypersetup{colorlinks=true,breaklinks=true, allcolors=black}

\usepackage{caption} 
\begin{document}

%\title{Characterizing player behavior in free-to-play games with time series 
%clustering techniques}
%\title{Clustering %synchronized 
%time series of player activity in free-to-play games}
\title{Discovering Playing Patterns: \\Time Series Clustering of Free-To-Play 
Game Data}


\author{\IEEEauthorblockN{ 
Alain Saas, Anna Guitart and \'{A}frica Peri\'a\~{n}ez}
\IEEEauthorblockA{Silicon Studio\\
1-21-3 Ebisu Shibuya-ku, Tokyo, Japan\\
Email: alain.saas@siliconstudio.co.jp, anna.guitart@siliconstudio.co.jp \\ 
africa.perianez@siliconstudio.co.jp}}

\maketitle

\begin{abstract}
The classification of time series data is a challenge common to all data-driven 
fields. However, there is no agreement about which are the most efficient techniques 
to group unlabeled time-ordered data. This is because a successful 
classification of time series patterns depends on the goal and the domain of 
interest, i.e. it is application-dependent.

In this article, we study free-to-play game data.
In this domain, clustering similar time series information is increasingly 
important due to the large amount of data collected by current mobile and web 
applications.
We evaluate which methods cluster accurately time series of mobile games,
focusing on player behavior data. We identify and validate several aspects of 
the clustering: the similarity measures and the representation techniques to reduce the high dimensionality of time series. As a robustness test, we compare various temporal datasets of player activity from two free-to-play video-games.

With these techniques we extract temporal patterns of player behavior relevant 
for the evaluation of game events and game-business diagnosis. Our experiments 
provide intuitive visualizations to validate the results of the clustering and to 
determine the optimal number of clusters. Additionally, we assess the common 
characteristics of the players belonging to the same group. This study allows us to improve the understanding of player dynamics and churn behavior.

\end{abstract}

% \fi
\IEEEpeerreviewmaketitle

\section{Introduction}

In the past years, free-to-play (F2P) has emerged as the dominant monetization model 
of games on mobile platforms \cite{fields2014mobile,annieidc}.
F2P games are offered for free, and monetized by charging for in-game content 
through in-app purchases, with player retention being key to a successful 
monetization.
The always-connected nature of mobile devices allows to constantly collect 
data about player behavior in the game. These data are used to guide design 
decisions for updates and release of additional content to maintain players' 
interest, sometimes in the form of periodic events giving access to new game 
content for a limited period of time \cite{xingwang}.

This study is motivated by the idea that the automatic clustering of time series of 
player behavior can lead to a better understanding of player engagement.
With daily active user bases ranging from thousands to millions of players, a game 
developer cannot know how every player reacts to a game or content update. At 
best, she can visualize averages of manually defined segments 
\cite{Segmentation}.
In this paper, we show that we can automatically cluster and visualize the main 
trends in player behavior and that we can determine differentiating 
characteristics of players belonging to different clusters.
We also consider the evolution of players after the end of the time series studied,
and we investigate the use of this clustering as a feature 
addressing temporal dynamics for further supervised learning applications, e.g. 
a churn prediction model.

Previous efforts on clustering game data appear in 
\cite{bauckhage2015clustering, drachen2012guns, drachen2014comparison, 
SifaBD14,DrachenSBT12}, with the common goal of extracting player pattern behavior. 
However, the focus of these studies is non-time-oriented data. On the other hand, 
in the work  presented by \cite{Menendez2014}, a clustering of time series is 
performed, but the measurements are obtained from PC games, not from F2P game data 
which allow a robust behavioral analysis.

The aim of the present paper is to identify similar patterns in unlabeled 
temporal datasets of player activity in F2P games. In order to discover 
natural groups of players, based on their behavior and interaction with the game, we 
apply diverse clustering techniques which focus on maximizing the dissimilarity 
between different clusters and maximizing the homogeneity within the 
groups. 

To the best of our knowledge this is the first article that applies unsupervised 
learning techniques to cluster time series of player behavior from F2P games. 
We have successfully extracted relevant user patterns from two F2P games: \emph{Age of 
Ishtaria} and \emph{Grand Sphere}, which helps us to examine quickly the player 
activity, allowing a visual game diagnosis and an intuitive evaluation of the 
weekly-based game events.

The games chosen for this study are representative of the most played F2P mobile 
social role-playing games in Japan and they have also been successful worldwide, reaching several millions of players.

\section{Clustering Time Series of Game Data} 
\label{clusters}
Time series consist of sequential observations collected and ordered over time. 
Nowadays, almost every application, web or mobile based, produces a massive 
amount of time series data. The goal of unsupervised time series learning, e.g. 
clustering methods, is to discover hidden patterns in time ordered data. 

Clustering time series data has received high attention over the last two 
decades \cite{Fu:2011}, starting with the seminal work of \cite{Agrawal:1993} in 
1993. It has faced many challenges \cite{Esling:2012}, among which one of the most 
important is probably the high dimensionality level that time series contains and 
therefore the difficulty of defining a similarity measure, i.e. the {\em 
distance} between series, in order to classify close patterns in the same group. 
Working with {\em raw} time series is computationally expensive and technically 
complicated. So as to cluster them efficiently, their complexity must be 
reduced through {\em representation} techniques \cite{Ding:2008}, trying to 
maintain the  characteristic features of the data. Both the dimensionality 
reduction and the similarity distance definition are obviously application-dependent.

Furthermore, with the fast increase of digital data, the clustering algorithms 
must be ready to deal with {\em Big Data} challenges \cite{Xi:2006}, e.g. a vast 
volume of data to be processed with high efficiency and speed, sometimes even in 
real time. The literature on time series clustering is very extensive. For a 
comprehensive review about time series similarity search methods, check 
\cite{liao2005clustering, Wang2006, Keogh:2003,Keogh:2006}.

In this Section we review the methods applied in the present paper to cluster 
the time series of player behavior. We briefly explain separately the 
representation methods and similarity measures used to evaluate the clustering 
results.

\subsection{Similarity Measures}
\label{sim}
Given a time series, defined as a sequence such as 
An\begin{equation}
X_n = (x_1, x_2, \cdots, x_N), 
\end{equation}
where $x$ are the observations measured at different times $n$, we need to 
determine the level of  {\em similarity}/{\em dissimilarity} (i.e. 
agreement/discrepancy) between a pair of them in order to cluster a sample of 
$K$ time series. 

Traditional distance computations, such as the Euclidean distance, can produce 
interesting results. However, in the case of \emph{time series}, notions of 
distance need not to be confined to this simple geometric paradigm.

There are several ways to measure the dissimilarity between pairs of time 
series.
This paper aims to cluster player profiles from \emph{Age of Ishtaria} and 
\emph{Grand Sphere} games based on their in-game behavior, hence dissimilarity 
measures were chosen according to this business target. 

We are interested in the so-called \emph{model-free} measures 
\cite{montero2014tsclust}.
A naive \emph{model-free} approach is to treat each series as an $n$-dimensional 
vector, and to calculate a shape-based geometric distance measure (without taking 
into account the absolute value of the time series selected). Such measures are 
the focus of our work, as we are interested in the \emph{shape} pattern behavior 
(geometric comparison) rather than the magnitude of the time series.   

Among the dissimilarity methods tested, those which provide the most robust 
results to classify time series are: Euclidean distance, Correlation (COR), Raw 
Values and Temporal Correlation (CORT) and Dynamic Time Warping (DTW).
In addition, a {\em complexity-based}  approach \cite{montero2014tsclust} called 
Complexity  Invariant Distance (CID) \cite{batista2014cid} is applied. With this 
method, instead of focusing on the shape of the series, we expect to group 
profiles from a different perspective, taking into account the degree of 
variability over time.  

Some other measures were also evaluated, e.g autocorrelation-based dissimilarity, 
Frechet distance measure, periodogram-based dissimilarity, among others (a 
review about these techniques can be found in \cite{montero2014tsclust}). 
However, these methods did not output as satisfactory results as the ones mentioned in the previous paragraph. The selected measures of interest are defined below, considering two time 
series 
$X$,$Y$ of size $T$ (which is the temporal dimension).
%%%%%%%%%%%%%%%%%%%%%%%%%%%%%%%%%%%%%%%%%%%%%%%%%%%%%%%%%%%%%%%%%%%%%%%%%%%%%%%%
%%%%%%%%%%%%
% DISSIMILARITY MEASURES EXPLANATION
%\begin{itemize}
%%%%%%%%%%%%%%%%%%%%%%%%%%%%%%%%%%%%%%%%%%%%
%%	
   \subsubsection{Dynamic Time Warping (DTW)}
DTW is a {\em non-linear} similarity measure obtained by minimizing the distance 
between two time series \cite{Bemdt94usingdynamic}. This method permits to group 
together series that have similar shape but out of phase 
\cite{ralanamahatana2005mining}. Figure \ref{compSim} shows in the left bottom 
panel how DTW aligns series with delayed but similar patterns.
   
DTW distance can be expressed as
   \begin{equation}
   DTW(X,Y) = \min\limits_{r\in M} \left( \sum_{m=1}^{M} |x_{im}-y_{jm}| 
\right),
   \end{equation}	
where the path element $r=(i,j)$ represents the relationship between the two series.
The goal is to minimize the time warping path $r$ so that summing its $M$ components gives the lowest measure of minimum cumulative distance between the time series. DTW searches for the best alignment between $X$ and $Y$, computing the minimum 
distance between the points $x_i$ and $y_j$.

%%%%%%%%%%%%%%%%%%%%%%%%%%%%%%%%%%%%%%%%%%%%%%

%%%%%%%%%%%%%%%%%%%%%%%%%%%%%%%%%%%%%%%%%%%%%%		
	\subsubsection{Correlation-based measure (COR)}
	It performs dissimilarities based on the estimated Pearson's correlation 
of two given time series. The COR computation can be expressed as
	\begin{equation}
	\begin{split}
	COR(X,Y) = \ \ \ \  \ \ \ \  \ \ \ \  \ \ \ \  \ \ \ \  \ \ \ \  \ \ \ \ 
 \ \ \ \  \ \ \ \  \ \ \\  \frac{\sum_{n=1}^{N}(x_n - \bar{X})(y_n - 
\bar{Y})}{\sqrt{\sum_{n=1}^{N}(x_n-\bar{X})^2}\sqrt{\sum_{n=1}^{N}(y_n-\bar{Y}
)^2}}.
	\end{split}
	\end{equation}

%%%%%%%%%%%%%%%%%%%%%%%%%%%%%%%%%%%%%%%%%%%%%%		    				
		
	\subsubsection{Temporal Correlation and Raw Values Behaviors measure 
(CORT)}
	It computes an adaptive index between two time series that 
covers both dissimilarity on raw values and dissimilarity on temporal 
correlation behaviors.
It can be written as
   \begin{equation}
   \begin{split}
   CORT(X,Y) = \ \ \ \  \ \ \ \  \ \ \ \  \ \ \ \  \ \ \ \  \ \ \ \  \ \ \ \  \ 
\ \ \  \ \ \ \  \ \ \\ \frac{\sum_{n=1}^{N-1}(x_{n+1} - x_n)(y_{n+1} - 
y_n)}{\sqrt{\sum_{n=1}^{N-1}(x_{n+1}-x_n)^2}\sqrt{\sum_{n=1}^{N-1}(y_{n+1}
-y_n)^2}}.
   \end{split}
   \end{equation}
   
%    This similarity measure takes values within the range $\in [-1,1]$. Hence, if 
% both series are similar the \emph{CORT} value is close to 1, on the other hand 
% if both series are similar but in opposite direction \emph{CORT} will be -1. 
% \emph{CORT} is 0 when there is no monotonicity between the time series $X$ and 
% $Y$.\\
   
%%%%%%%%%%%%%%%%%%%%%%%%%%%%%%%%%%%%%%%%%%%%%%	
    \subsubsection{Complexity-Invariant Distance measure (CID)}
CID computes the similarity measure based on the Euclidean distance but 
corrected by the complexity estimation of the series \cite{batista2014cid}. CID is 
written as
		\begin{equation}
		CID(X,Y) = dist(X,Y) \cdot CF(X,Y), 
		\end{equation}
		with $CF$ being the complexity correction factor defined by
		\begin{equation}
		CF(X,Y) = \frac{max(CE(X),CE(Y))}{min(CE(X),CE(Y))}.
		\end{equation}
		And $CE(\cdot)$ corresponds to the complexity estimations of a 
time series of length $N$, given by
		\begin{equation}
		CE(X) = \sqrt{\sum_{n=1}^{N-1}(x_n-x_{n+1})^2} .
		\end{equation}		

%%%%%%%%%%%%%%%%%%%%%%%%%%%%%%%%%%%%%%%%%%%%%%

\subsection{Representation Methods}

Time series data collected from F2P games are high dimensional objects. 
In order to reduce their complexity and make the comparison feasible,
it is convenient to perform a dimensional reduction transformation beforehand. There are several ways to 
reduce the data from $n$-dimensions to $N$-dimensions, and depending on the 
application domain, there are techniques more suitable than others. We focus on function approximation procedures to simplify the time series objects we aim to cluster. In the 
following subsections, we briefly summarize the most successful procedures for 
the video-game data tested in the present study. 

\subsubsection{Discrete Wavelet Transform (DWT)}
This method uses a wavelet decomposition to approximate the actual series. A 
wavelet is a function used to approximate the target time series by means of
superposition of several (wavelet) functions. A {\em wavelet} object provides 
information about variations of the time series locally, as it can be shown in 
Figure \ref{compSim} in the right lower panel. DWT assigns a coefficient to each {\em 
wavelet} component and the distance is computed between the wavelet-approximated 
time series. 
	\begin{figure}[t!]       
		\includegraphics[height=4cm]{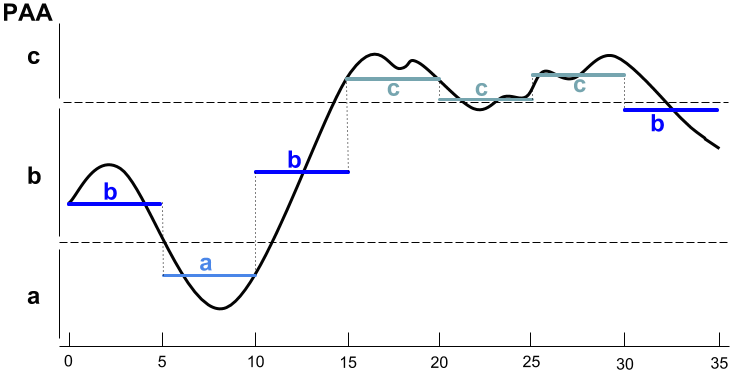}
        \caption{Illustration of SAX representation method 
(dimensionality reduction of time series), performed with the following parameter values: $w=7$ and $\alpha=3$.}
        \label{SAX}
	\end{figure}
    
\subsubsection{Symbolic Aggregate Approximation related functions measure (SAX)}
SAX is a symbolic representation to simplify continuous time series  \cite{lin2007experiencing}. The series is discretized and divided into sequential frames of equal size. Firstly, the series is divided in \emph{w} set intervals and it is represented by its corresponding mean Piece wise Aggregate Approximation (PAA) dimensional reduction.
Afterwards SAX is represented by a subset of alphabet letters of size $\alpha$ 
where $\alpha=(l_1,...,l_{\alpha})$ and the transformed series 
$\hat{X}=(\hat{X}_1,...,\hat{X}_{\alpha})$ is computed by determining 
equal-sized zones under a Gaussian distribution.
The distance is then computed between the approximated time series. Figure 
\ref{SAX} depicts a schematic view of the SAX dimensionality reduction method.

\begin{figure}[t!]
\centering
\includegraphics[width=\columnwidth]{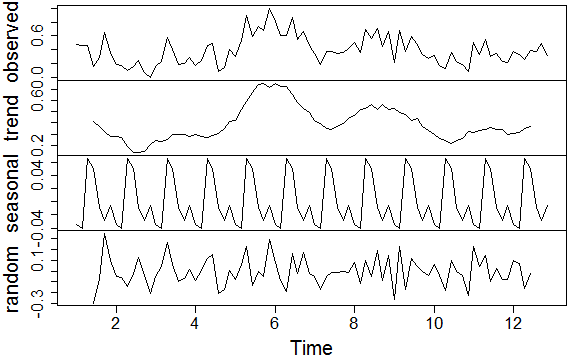}
\caption{Time Series Decomposition used for dimension reduction. Original series 
in the upper panel, trend extraction in the second plot, the seasonal and the 
random (residual fluctuations) components shown in the third and fourth 
panel, respectively.}
\label{size}
\end{figure}

\subsubsection{Trend Extraction}

A time series is composed of different elements such as seasonal components, 
medium or long-term trend, cyclical movement (repeated pattern but non-periodic) 
or irregular fluctuations (also known as residuals).
Depending on the clustering application, some of these components can be 
representative of the characteristics of the \emph{raw} time series.

As we are interested in characterizing players by their behavior, we focus on 
the trend component as it provides essential information for this purpose. 
Seasonality behavior or irregular data are not the center of our attention, the trend rather reflects significant information about player's interests. As it 
is mentioned in \cite{AlexandrovTrend} ``The trend of a time series
is considered as a smooth additive component that contains information about global change".
The method used in this work to extract the trend is the moving average filtering. 

\subsection{Hierarchical clustering}
In this subsection, we describe the methods to create the nested partitions in order to classify the total number of time series. 

Hierarchical clustering creates homogeneous partitions of data according to 
their level of dissimilarity, maximizing the difference between clusters 
\cite{murtagh2012algorithms,murtagh2011ward}. The clustering growing method can 
be increasing (agglomerative clustering or bottom-up) or decreasing (divisive 
clustering or top-down) at each step.
It is normally represented by dendrograms that show the clustering levels in a 
tree-based graph. Figure \ref{plotDendrogram} illustrates with a dendrogram the 
hierarchical clustering performed to classify player behavior of {\em Age of 
Ishtaria}.

The method selected to cluster the datasets studied in the current paper is {\em 
agglomerative clustering}.
There are different methods of agglomerative clustering 
\cite{murtagh2012algorithms}, but the one used for our analysis is the so-called 
\emph{Ward method} which is a minimum variance technique. In Ward, the distance 
between two clusters is defined as the deviance between them. The clusters that are merged in the same group are the ones that lead to a minimum increase in the 
total within-cluster variance (calculated from the dissimilarity measure 
selected between the time series) \cite{murtagh2011ward}. This method is used to obtain the 
results presented in the Section \ref{results}, as our goal is to obtain a low 
variance within the clusters.

\begin{figure}[t!]
\centering
\includegraphics[width=0.49\columnwidth]{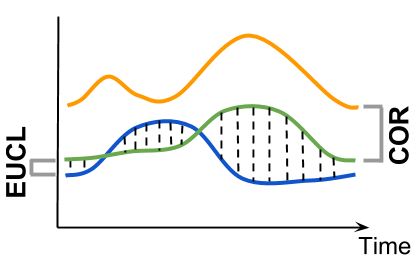}
\includegraphics[width=0.49\columnwidth]{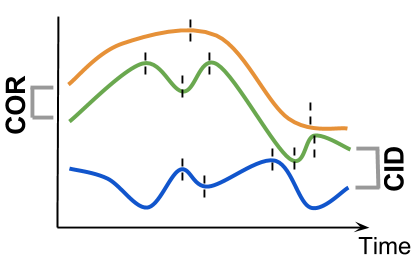}\\
\includegraphics[width=0.49\columnwidth]{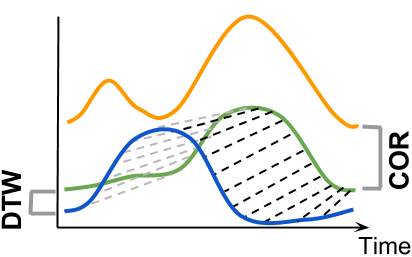}
\includegraphics[width=0.49\columnwidth]{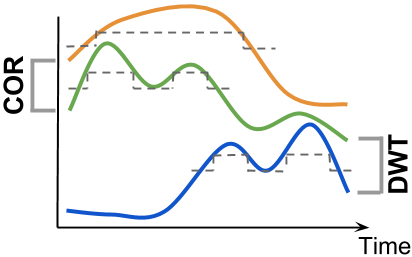}
\caption{Illustrations of the difference between time series clustering results 
obtained using Euclidean (EUCL) and Correlation dissimilarity measure (COR) 
(left upper panel), Dynamic Time Warping (DTW) in the upper right panel, 
Complexity Invariant Distance (CID) in the left lower panel, and Discrete 
Wavelet Transform (DWT) in the right lower panel.}
\label{compSim}
\end{figure}
%%%%%%%%%%%%%%%%%%%%%%%%%%%%%%%%%%%%%%%%%%%%%%
\section{Comparison of Clustering Methods}
The selection of an adequate technique to cluster time series depends on the 
application and business interest. All the methods reviewed in Section \ref{clusters} were tested to cluster the time series of game data. We want to classify players by pattern-shape, 
without giving too much attention to small fluctuations or to the total magnitude of the 
time series. Based on this aim we can conclude:

\begin{itemize}

\item DTW works particularly well to group similar player profiles with a shift 
on the time axis. DTW also groups 
together similar patterns but at different scale. However, as we are 
interested in evaluating the impact of game events on player activity, we 
rather focus on clustering synchronized profiles. Therefore, this is not the most 
suitable tool to measure the distance between time series for the purpose of the current study.

\item In the DWT method of dimensionality reduction, the wavelets define the 
frequency of the series, which sometimes does not fit with the weekly 
seasonality we want to study. 

\item SAX representation method could be a useful tool for our problem as we 
are interested in identifying the pattern behavior, and not detailed aspects of the 
time series. However, the manual tuning of the two parameters $w$ and $a$ can be 
a drawback, although it is easily done once we introduce apriori information 
about the seasonality of the time series (the weekly events in our case). We 
hoped that the dimensionality reduction offered by SAX would allow us to cluster 
longer time series (2 months or more) but we did not obtain conclusive 
results.

\item COR is a promising method for our goal. It groups similar geometric and 
synchronous profiles. As a drawback, COR seems to be sensitive to noise data and 
outliers (which are present in our datasets).

\item CORT is similar to COR, but we ultimately obtained the most convincing 
results with the second.

\item COR+trend is the combination of COR and trend extraction, which addresses 
COR's sensitivity to noise. This method allows us to obtain the best 
results for non-sparse time series (such as the time series of time played). 
Indeed, the trend extraction does not work well with time series containing many 
zero values (such as the time series of in-app purchases).

\item CID groups series that have similar complexity patterns.
This method performs poorly in classifying similar geometric profiles, which is what we do successfully
with COR+trend applied to the time series of time played. However, it is 
the method that provides the best results when it comes to classifying time 
series containing large amount of zero values (such as the time series of 
purchases).
\end{itemize}
Figure \ref{compSim} shows an intuitive comparison between similarity measures 
and representation methods to help to understand the difference between different
techniques.  The similarity methods and representation techniques 
described in this paper were tested to obtain the results presented in Section \ref{results}.

\subsection{Evaluation Metrics of Clustering results}
\label{validation}
The validation of the clustering methods is a challenging task, as we do not 
have any \emph{truth} we can rely on to compare the accuracy of the classification, contrary to 
the supervised learning models. Several techniques to evaluate the adequacy of 
the similarity measures and representation methods to cluster time series were 
tested, among them:
Dunn and average of Silhouette width \cite{DunnSilhouette}, Normalized Hubert's statistic 
\cite{halkidi2001clustering} and Entropy \cite{meilua2007comparing}. However, due 
to the difficulty of the task and the high complexity of time series objects, the results were not satisfactory. We used several 
kinds of visualization techniques to validate the clustering results and to 
determine the optimal number of clusters.

%% TABLE OF CLUSTERING DESCRIPTION
\begin{table*}[!ht] \small
	\centering
     \caption{Summary description of the clustering results with \emph{Age of 
Ishtaria} and \emph{Grand Sphere} data.}
	\begin{tabular}{lcccccc} \hline
	\emph{Clustering result name} & \emph{Game} & \emph{Data} & 
\emph{Technique} &  \emph{Clusters}  & \emph{Start date} & \emph{End date} \\ 
\hline
Age of Ishtaria time clustering & Age of Ishtaria & Time played per day & COR+trend & 8 
& 11-Jun-2015 & 1-Jul-2015 \\
Age of Ishtaria spending clustering & Age of Ishtaria & Spending per day & CID & 5 & 
30-Jan-2015 & 19-Feb-2015\\
Grand Sphere time clustering & Grand Sphere & Time played per day & COR+trend & 
8 & 11-Sept-2015 & 1-Oct-2015\\ \hline
	\end{tabular}
	\label{tableDatasets}
\end{table*}

\section{Datasets}
\label{Data}
\subsection{Data Source}

Our data come from the games {\em Age of Ishtaria} and {\em Grand Sphere} by 
Silicon Studio.

% Our study focuses on time series of variables measured for each individual user. 
% The frequency of our time series is daily. The study is performed on a weekly 
% basis since there are weekly game events influencing player behavior. The 
% starting date of our time series is synchronized with the starting date of the 
% weekly events.

We worked with time series of the following variables that are game independent 
and can be measured in all free-to-play games. These variables are measured per user and per day.

\begin{itemize}
	\item \emph{Time}: The amount of time spent in the game
	\item \emph{Sessions}: The total number of playing sessions
	\item \emph{Actions}:  The total number of actions performed
	\item \emph{Purchase}: The total amount of in-app purchases
\end{itemize}

Time, sessions and actions time series are highly correlated and produce very 
similar results. Thus, for the purpose of our study, we focus on the Time 
variable, which has a lower measurement error.

Purchases time series are different from the others because they are sparse (they contain many 
zero values), as the majority of the paying users do not complete an in-app purchase 
every day.

\subsection{Time Series Studied}
The frequency of our time series is daily. We study them on a weekly basis since 
there are weekly game events influencing player behavior. The studied period 
$P$, therefore contains $N_{week}$ $\times $ 7 values, with $N_{week} = 3$ being the number of weeks selected for this study. We synchronize the 
starting date $P_{start}$ and the ending date $P_{end}$ of our time series with the 
starting date of the game events.

In order to avoid a bias due to the partial absence of data from players who join 
or leave the game during $P$, we only consider data from players who installed 
the game before $P_{start}$ and who are still active after $P_{end}$.

For the final results of clustering the time series of Time played, we 
consider only data from the users who played at least 6 days per week. There 
are two reasons for this choice. Firstly, from a free-to-play game developer 
perspective, we are interested in the most active players. Secondly, the 
clustering technique that allowed us to obtain the best results for this 
clustering performs poorly with sparse time series.

For the final results of clustering the times series of Purchases, which 
are mostly sparse, except for the very top spenders, we considered the players 
who did at least one purchase during $P$. Due to the sparse nature of these time 
series, we then obtain the best results using a different clustering technique.

Finally, we take random samples of 1000 time series respecting the conditions above for our experiments.

\begin{figure}[!t]
    \center
    \includegraphics[width=0.49\textwidth]{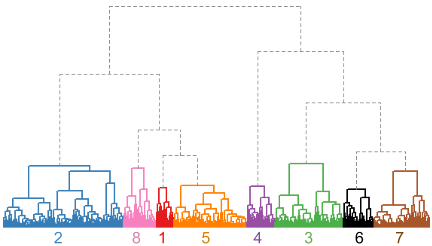}
    \caption{Hierarchical Clustering represented by a dendrogram of {\em Age 
of Ishtaria} time-played data. The Age of Ishtaria time clustering is performed with COR similarity measure and trend 
extraction as representation method.}
	\label{plotDendrogram}
\end{figure}

\section{Results}
\label{results}
In this section, we present the results of the clustering experiments summarized 
in Table \ref{tableDatasets}. For each experiment, we call event A, B and C the 
game events released respectively on week 1, 2 and 3. This is a naming convention 
independent of the content of the events.

\begin{figure*}[!t]
    \center
    \includegraphics[width=\textwidth]{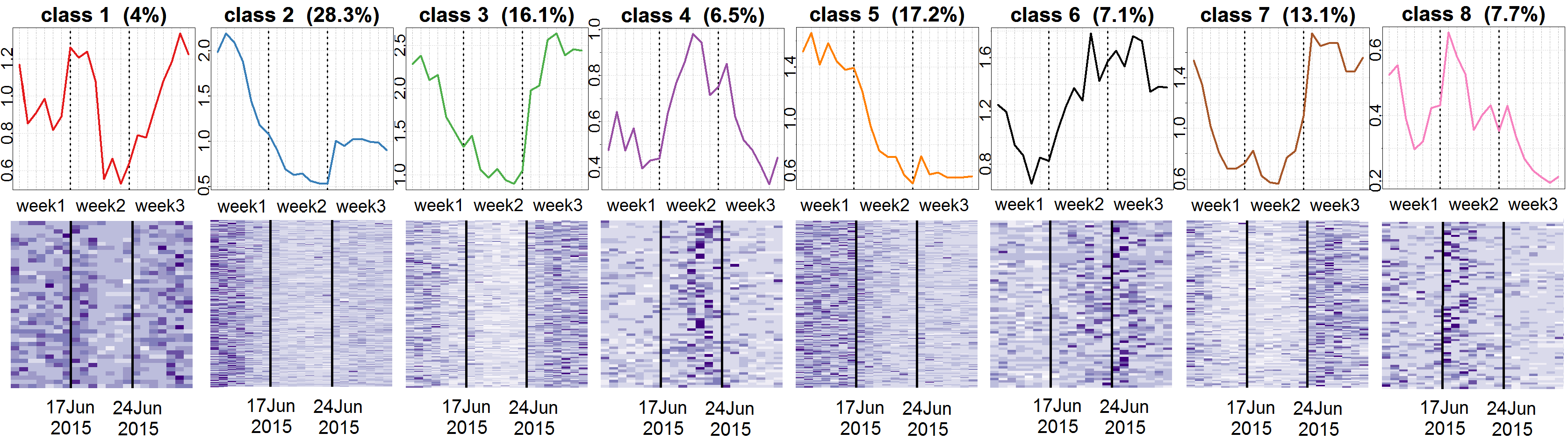}
    \caption{Mean of the time series and heatmap for each cluster from Age of Ishtaria time clustering (time played per day). Vertical lines delimiting the game events. Clustering performed with COR similarity measure and trend extraction.}
	\label{plotAllMeans}
\end{figure*}

\begin{figure*}[!h]
    \center
    \includegraphics[width=\textwidth]{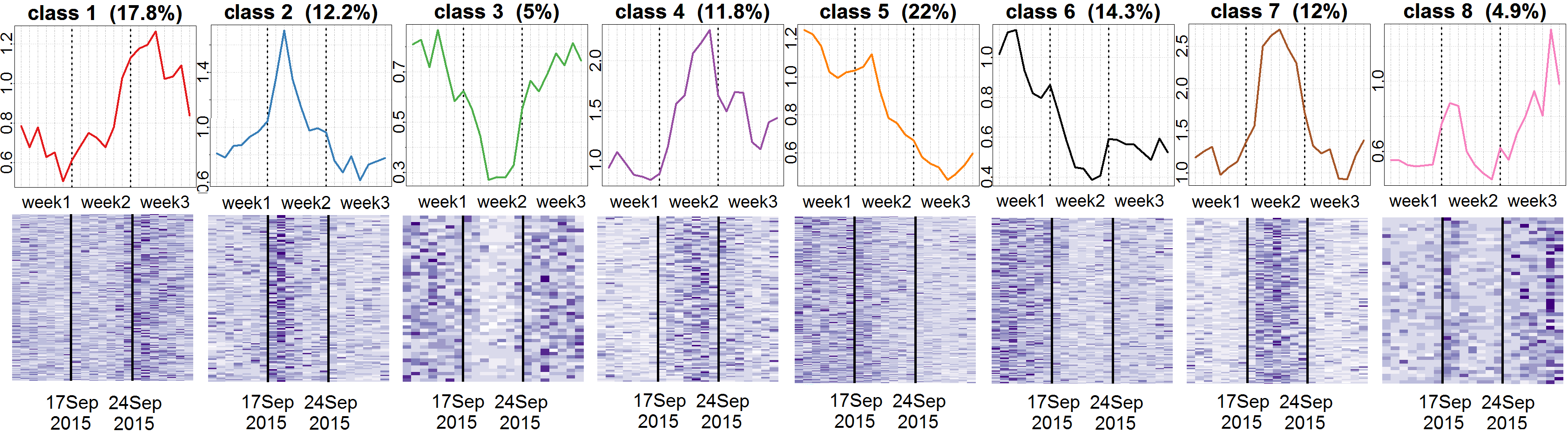}
    \caption{Mean of the time series and heatmap for each cluster from Grand Sphere time clustering (time played per day). Vertical lines delimiting the game events. Clustering performed with COR similarity measure and trend extraction.}
	\label{plotAllMeansGransphere}
\end{figure*}	

\begin{table*}[!ht] \small
	\centering
    \caption{Characteristics of the players at the starting date of the 
studied period (Age of Ishtaria time clustering)}
	\begin{tabular}{lcccccccc} \hline
\emph{variables}   & \emph{class 1}& \emph{class 2}& \emph{class 3}& \emph{class 
4}& \emph{class 5}& \emph{class 6} & \emph{class 7} &\emph{class 8}\\ \hline
number of players&40	&283	& 161&	65&	172&	71 & 131&	77\\
ratio PU&30.0\% &33.2\% &44.7\% &20.0\% &33.1\% &33.8\% &35.1\% &14.3\% \\ 
average	level&47&53&75 &35&51&49&58&31\\\hline
	\end{tabular}
	\label{tableResultsVariables}
\end{table*}

\begin{table*}[!t] \small
	\centering
    \caption{Cumulative churn ratio in the months following the clustering, 
after period $P$ (Age of Ishtaria time clustering)}
	\begin{tabular}{lcccccccc} \hline
\emph{churners ratio}   & \emph{class 1}& \emph{class 2}& \emph{class 3}& 
\emph{class 4}& \emph{class 5}& \emph{class 6} & \emph{class 7} &\emph{class 
8}\\ \hline
July     & 15.0\% & 11.7\% &4.3\% &15.4\% &19.2\% &18.3\% &5.3\% &22.1\% \\
August   & 27.5\% & 19.8\% &13.0\% &26.2\% &30.8\% &31.0\% &14.5\% &28.6\% \\ 
November & 45.0\% & 48.4\% &29.2\% &47.7\% &51.7\% &50.7\% &32.8\% &58.4\% 
\\\hline
	\end{tabular}
	\label{tableChurn}
\end{table*}

\begin{figure*}[!h]
    \center
    \includegraphics[width=\textwidth]{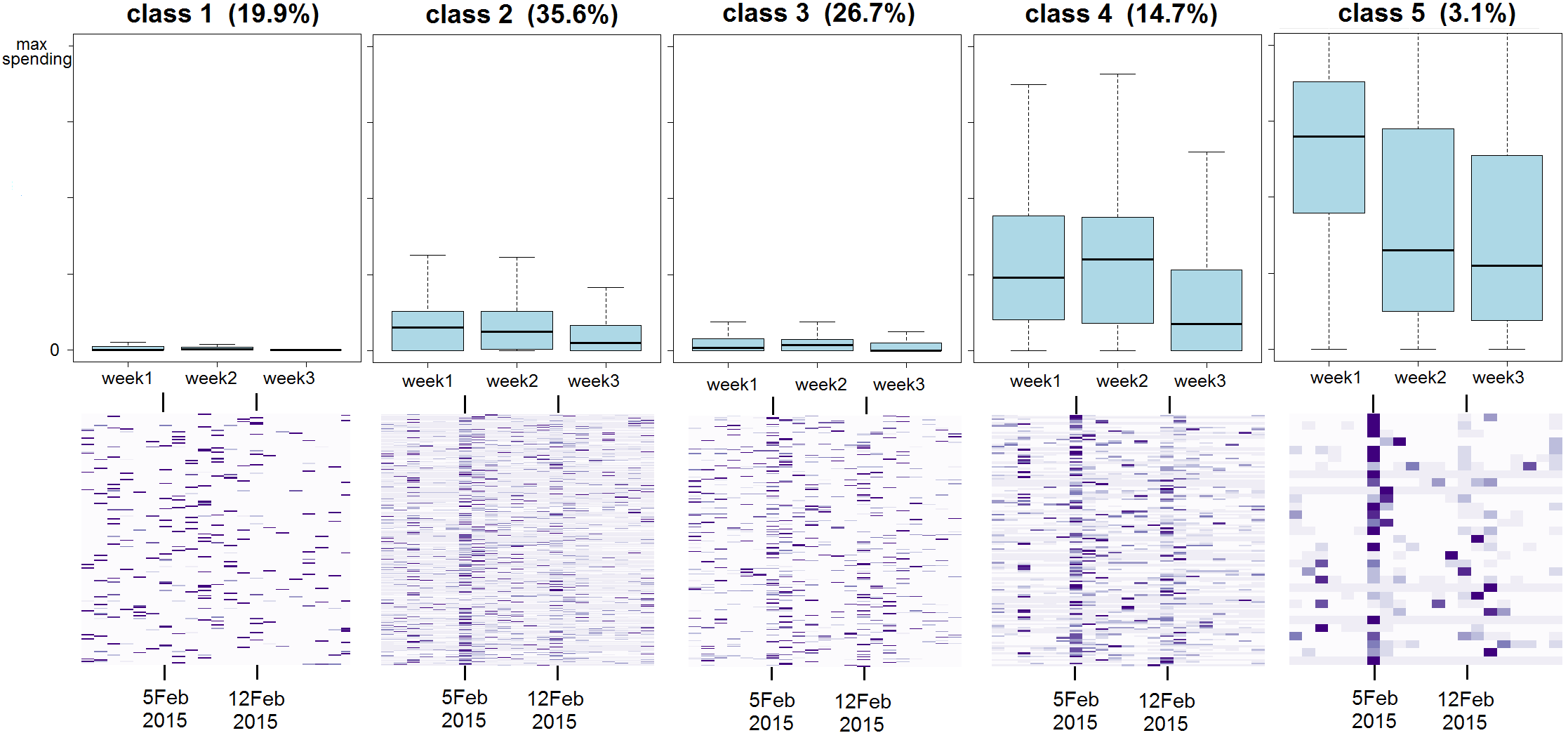}
    \caption{Clustering results from the visualization of purchase's time series from Age of Ishtaria  
data using CID similarity measure (Age of Ishtaria spending clustering). Box plots of the spending per player and per week in the upper 
panel. Corresponding heatmap for each cluster below. The dates on the $x$-axis 
delimiting the weekly game events.}
	\label{plotAllMeansIshtariaCID}
\end{figure*}	
\subsection{Clustering and Visualization}

\subsubsection{Clustering time series of time played} 

Figure \ref{plotAllMeans} visualizes the classification obtained by clustering time series of time 
played per day by Age of Ishtaria players, using the correlation dissimilarity 
measure on the trend extracted from the raw time series. We call this method 
\emph{COR+trend}.

For each cluster, we plot the mean of the time series and a heatmap containing 
all the time series (one time series for each player included in the sample).

Visualizing the mean allows to see the top trends in player behavior. For 
example, the activity of class 3 plunged for event B but spiked for event C, 
while class 4 followed an opposite pattern. Since the game events are usually
designed based on predefined game templates (i.e. they are reused throughout the
lifetime of the game), this analysis helps the game designer to better understand the
interest of several groups of players in different kinds of game events. This supports
future game event planning and improves the knowledge about the impact of the game events on the player activity.

Visualizing the time series of each cluster on a heatmap allows to quickly 
validate the quality of the clustering. Figure \ref{plotAllMeans} shows that the time 
series follow the same patterns within each cluster.

Heatmap along with a dendrogram visualization, represented in Figures 
\ref{plotDendrogram} and \ref{plotAllMeans}, proves to be a better tool than the statistical measures 
tested, mentioned in Section \ref{validation}, for choosing the optimal number of clusters. Using 
these tools we determine that the most optimal clustering is obtained with 
8 clusters.

We apply the same clustering technique on time series of time played by the 
players of Grand Sphere, and obtain similar results, as it can be checked in Figure
\ref{plotAllMeansGransphere}. This is a promising fact towards obtaining an 
adequate technique ready to cluster data from other F2P games.

\subsubsection{Clustering time series of purchases}
Figure \ref{plotAllMeansIshtariaCID} depicts the clusters obtained by clustering 
time series of purchases per day by Age of Ishtaria players, using the Complexity 
Invariant Distance (CID) on the raw time series.

For each cluster, we represent the distribution of the data
separated per week in a box-plot, and a heatmap containing all the time series.

Visualizing the time series of each cluster on a heatmap allows to distinguish 
different purchase patterns. For example, players from class 1 and class 3 
purchase sparsely while players from class 2 purchase nearly every day.

Since the scale of each heatmap is normalized separately to be able to visualize 
properly the full range of purchases on each heatmap, we can not compare the 
amount of the purchases between the clusters using only this visualization. And, 
contrary to the clustering of the time series of time played described above, 
the time series of purchases are mostly sparse, which makes it irrelevant to plot 
the mean of these time series. That is why we use the box-plot 
representation of the spending for each week, in order to visualize the 
difference of scale between the different groups. This additional plot allows 
us, for example, to see that class 5 contains very high spenders even if 
they have a relatively sparse purchase behavior like class 1 and 3.

\subsection{Extraction of Players Characteristics}
We are not only interested in clustering game users and discovering hidden patterns, 
but also want to analyze the characteristics they have in common.

After performing the clustering, we measure how 
players behave during the period $P$ by analyzing their 
characteristics on the start date $P_{start}$ of the time series.

Table \ref{tableChurn} reflects that class 3 contains players with the highest 
playing levels and also the highest ratio of paying users, while classes 4 and 8 contain 
the players with the lowest levels and the lowest ratios of paying users.

It is interesting to note that these clusters coincide with the ones already 
discussed earlier as reflecting an opposite interest in certain game events. 
With these two observations, we can conclude that event B was unpopular for 
advanced players and more popular for less advanced players, and that 
event C was more popular for advanced players and unpopular for less advanced 
players. A game planner visualizing this could conclude that she had better 
avoid triggering an event of event C's type soon after a user acquisition 
campaign, as it would likely be unpopular for the new coming less advanced
players just acquired.

This example shows that it is possible to extract differentiating player 
characteristics from the clustering we obtained.

\subsection{Churn behavior}
Player retention is of crucial importance in F2P games. Several models have been 
proposed to help to understand and predict the churn of players \cite{hadiji, 
runge2014churn, churnpredictionOurs}.

Based on the results obtained in Table \ref{tableResultsVariables} and Figure 
\ref{plotAllMeans}, we study the evolution of the players after the period of 
time $P$ covered by the time series, in order to see if there is a relation 
between their behavior during $P$ and after $P$.

Table \ref{tableResultsVariables} shows the churning rate 1, 2 and 5 months 
after the period $P$ for each cluster. We observe that class 3 and 7 have 
a significantly lower churning rate than class 4 and 8, being 3 to 4 times 
lower after 1 month and 1.5 to 2 times lower after 5 months.

According to this result, players have a different churn behavior following 
their profile classification performed during the period $P$.

Therefore, the use of the unsupervised classification of player profiles 
suggested in this article could be an interesting feature to address the temporal 
dynamics of players data for a churn supervised learning model. In 
\cite{runge2014churn} an alternative approach was proposed using a Hidden Markov 
Model. 

However, in order to use this predictor in a supervised model some changes need to 
be performed in the definition of the problem as we discussed in Section 
\ref{Data}. This comprehensive analysis is beyond the scope of this paper. For 
example, this would involve to cluster players based on their last weeks behavior, 
e.g. the time series starting date would be 3 weeks before the last day the players 
connected to the game instead of taking fixed dates as in the present work.

This time series classification would allow us to improve the understanding 
about the churn of players but, on the other hand, it would not provide 
information about game events reaction, which is a principal target of the 
current analysis.

\section{Summary and Conclusion}
In the present article, we have conducted a research about unsupervised 
clustering of time series data from two free-to-play games. We evaluate several 
similarity measures and representation methods to extract meaningful behavioral 
patterns of players. This allows us to assess the impact of weekly game events 
and discover hidden playing dynamics regarding purchases and time played per 
day. An appropriate characterization of time series allows us to find significant 
attributes in common among players belonging to the same group.  Ongoing and 
future work involve the application of the time series clustering results to 
churn prediction models and further analysis of the player profiles. 

\section{Software}
The analyses presented in Section \ref{results} were performed with the R 
version 3.2.3 for Windows, using the following packages from CRAN: 
\emph{TSclust} 1.2.3 \cite{montero2014tsclust}, \emph{timeSeries} 3022.101.2 
\cite{wuertz2009timeseries}, \emph{fpc} 2.1-10 \cite{hennigfpc}, \emph{Rmisc} 
1.5 \cite{hope2013rmisc}, \emph{reshape} 0.8.5 \cite{wickham8reshape}, 
\emph{ggplot2} 2.0.0 \cite{wickham2013implementation}. 

\section*{Acknowledgments}
% We thank our colleagues at Silicon Studio: Sovannrith Lay from our data science team, Hiroshi Okuno from the marketing team, and Takeshi Kimura, Tomomi Hamamura, Kotaro Narizawa, Yumi Kida from the game tea from the game team,m, for their help to collect the data and their support during this study.
We thank our colleagues Sovannrith Lay, Hiroshi Okuno, Takeshi Kimura, Tomomi Hamamura, Kotaro Narizawa and Yumi Kida for their help to collect the data and their support during this study. We also thank Thanh Tra Phan for the careful review of the article.

\bibliographystyle{IEEEtran}
\bibliography{main}

% that's all fol0ks
\end{document}